# MoVISA: Multi-Token Reasoning for Video Object Segmentation

**Ruining Zhao** **Ho Kei Cheng** **Alexander G. Schwing**
University of Illinois Urbana-Champaign
{ruining9, hokeikc2, aschwing}@illinois.edu

## Abstract

Recent advances in video object segmentation with Multimodal Large Language Model (MLLM) reasoning have demonstrated the effectiveness of using a single textual token (e.g., <SEG>) to predict segmentation masks across images and videos. However, we observe that this single-token strategy lacks the granularity required to precisely localize multiple objects across time in video segmentation tasks. To address this limitation, in this work, we develop "Multi-Token Reasoning for Video Object Segmentation" (MoVISA). MoVISA uses multiple segmentation tokens (e.g., <SEG0>, <SEG1>, ...) to represent an object across different frames. This design enables a more fine-grained alignment between language prompts and spatio-temporal mask predictions, enhancing both performance and interpretability. On the challenging MeViS, DAVIS17, ReVOS, and Ref-Youtube-VOS benchmarks, our model achieves a 13.2% J&F improvement on MeViS and a 8.4% J&F improvement on ReVOS. Code and models will be released.

## 1 Introduction

The integration of Multimodal Large Language Models (MLLMs) into visual understanding has led to remarkable results for perception tasks ranging from object detection to segmentation. Building on the alignment of vision and language through joint embeddings, recent work such as GPT-4 [OpenAI et al., 2024], MiniGPT-4 [Zhu et al., 2023], LLaVA [Liu et al., 2023], and mPLUG-Owl [Ye et al., 2024], demonstrated that MLLMs can reason in complex visual scenes when provided with high-quality visual inputs and well-aligned textual prompts. The resulting models extend the capability of Large Language Models (LLMs) by incorporating frozen or fine-tuned vision backbones and by projecting vision features into language space, enabling open-ended visual reasoning.

For referring object segmentation, LISA [Lai et al., 2024] introduced a pioneering approach for reasoning-based image segmentation by leveraging a single special token, <SEG>, to represent the spatial region referred to in natural language. The model generates segmentation masks by predicting a <SEG> token and decoding the hidden features of <SEG> in the final layer. Building on this concept, recent works such as VISA [Yan et al., 2024] and Video-LLaVA [Lin et al., 2024] have extended this single-token paradigm to the video domain. These approaches typically sample a limited number of frames and apply the same <SEG> token across all frames, enabling the model to generate object masks based on temporally sparse visual observations. However, these single-token methods face challenges when dealing with long or complex videos where a single representation may not suffice to capture the diverse temporal appearances of an object. To address this, Sa2VA [Yuan et al., 2025] introduces a more advanced framework for reasoning-based video segmentation by integrating the recently proposed SAM2 architecture [Ravi et al., 2025] as its segmentation backbone for temporally propagating frame-level segmentation over time. Despite the advances of the memory mechanism of SAM2, the single segmentation token design still suffers from fundamental limitations to represent local information at each frame. In many real-world scenarios, such as tracking a moving object or resolving temporally-anchored expressions (e.g., "Please segment the horse running around"), it becomes increasingly difficult for a single token to represent diverse spatial manifestations of the object across time. The reliance on a single global `<SEG>` embedding introduces an information bottleneck in long-range tracking or cluttered scenes.

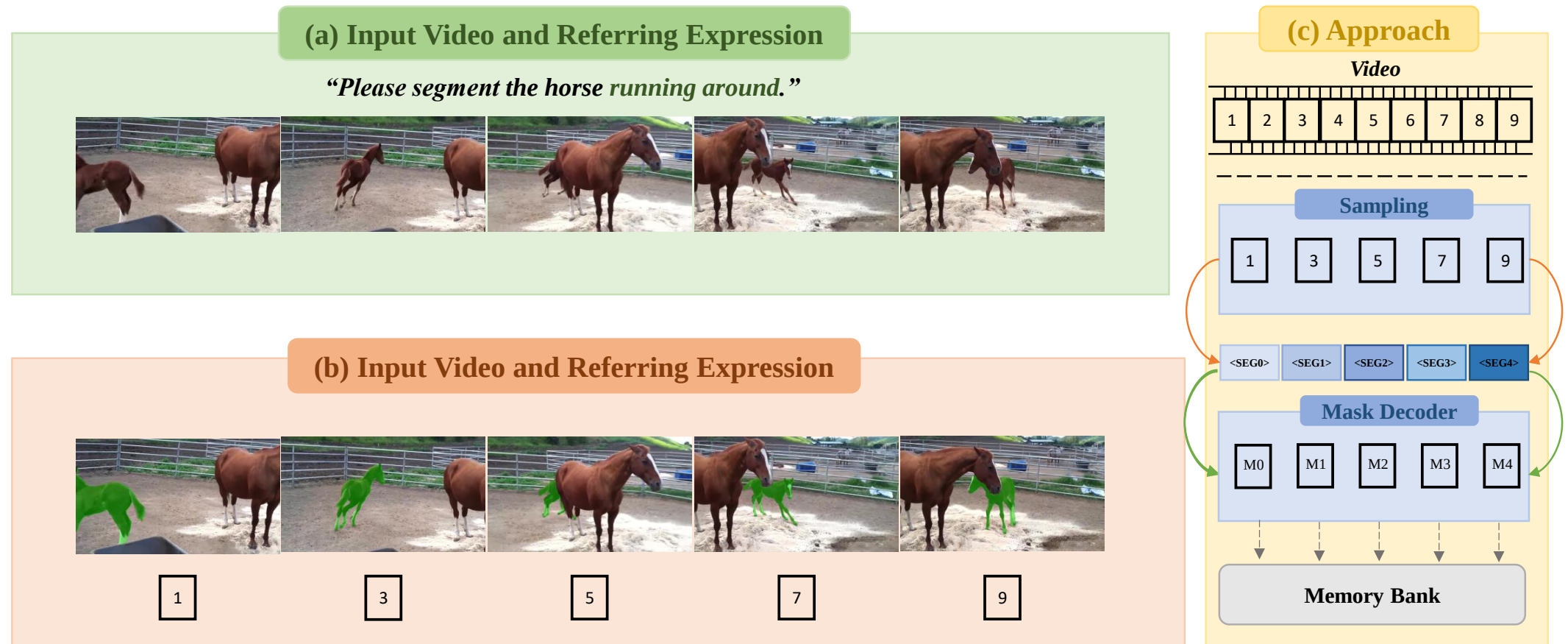

Figure 1: Overview of MoVISA. Given an input video and a referring expression, MoVISA samples a subset of key frames and generates one segmentation token per frame. The segmentation tokens are processed by a mask decoder to generate the corresponding masks, which are then added to a memory bank for mask propagation to non-key frames.

To alleviate this bottleneck, we present Multi-Token Reasoning for Video Object Segmentation (MoVISA). It introduces fine-grained, object-frame aligned `<SEG_n,t>` tokens for each object $n$ at frame $t$, as illustrated in Figure 1. Instead of collapsing all spatial-temporal reasoning into a single representation, our approach allows the model to reason about each frame-object pair independently to capture subtle variations across the video. Each `<SEG_n,t>` token is projected into a spatial segmentation mask via a modified SAM2.

Our contributions are summarized as follows:

- We propose multi-token prompting, introducing explicit, object-frame aligned segmentation tokens into vision-language prompts, and enabling precise spatio-temporal segmentation.
- We integrate this token design into a unified, end-to-end trainable framework with SAM2 and MLLM backbones, without requiring external VOS or heuristic post-processing modules.
- We demonstrate that our method consistently outperforms single-token baselines on Ref-YoutubeVOS, MeViS, and ReVOS.

## 2 Related Work

### 2.1 Multi-Modal Large Language Models

Large Language Models (LLMs) have demonstrated exceptional capabilities in language understanding and reasoning. This success has spurred the development of Multi-Modal Large Language Models (MLLMs) that bridge visual and linguistic information. For static images, early works like Flamingo [Alayrac et al., 2022] adopt cross-attention to condition LLMs on visual tokens, enabling in-context visual learning. Dual-encoder approaches such as BLIP-2 [Li et al., 2023b] and mPLUG-OWL [Ye et al., 2024] project image embeddings into the LLM token space for downstream reasoning tasks. Instruction-tuned models like Otter [Li et al., 2023a], LLaVA [Liu et al., 2023], and MiniGPT-4 [Zhu et al., 2023] further enhance alignment by adapting the models on large-scale vision-language corpora.

Extending MLLMs from images to video data introduces significant challenges, including temporal modeling and long input sequences. Video-LLaMA [Li et al., 2024] extends BLIP-2 by extracting frame-wise embeddings and injecting them into LLMs. Meanwhile, Video-ChatGPT [Maaz et al., 2024] employs spatio-temporal pooling to compress video inputs, albeit at the cost of losing fine-grained detail. LLaMA-VID [Li et al., 2024] mitigates the token explosion problem by encoding each frame into just two tokens, allowing longer sequences to be handled but sacrificing spatial precision. To overcome the issue of spatial-temporal grounding brought by image-based MLLM, Sa2VA [Yuan et al., 2025] integrates SAM2 [Kirillov et al., 2023], a fully differentiable segmentation model with a memory bank and temporal aggregation. This allows the model to propagate object representations over time and support pixel-level video reasoning. Our method builds on top of these advancements in MLLMs and further proposes using multiple segmentation tokens to better integrate MLLMs with foundational segmentation models like SAM2.

### 2.2 Video Object Segmentation

Video object segmentation consists of referring segmentation and reasoning segmentation. Referring segmentation localizes objects described by natural language, generating spatial masks in images or temporally consistent masks in videos. Earlier works [Yu et al., 2018, Khoreva et al., 2018, Ding et al., 2023] focused on multi-modal fusion to align object features and linguistic descriptions. Transformer-based methods like UniPercept [Yan et al., 2023] unify segmentation and tracking under a shared attention-based framework to improve temporal consistency.

Recent trends leverage MLLMs to model high-level reasoning for segmentation. LISA [Lai et al., 2024] incorporates a special token `<SEG>` and uses its hidden embedding for mask decoding, which effectively binds language and vision space. VISA [Yan et al., 2024] and Video-LLaVA [Lin et al., 2024] extended this to videos using shared `<SEG>` tokens across sampled frames. Sa2VA [Yuan et al., 2025] leverages SAM2 [Kirillov et al., 2023], which supports memory-based feature propagation and fully differentiable segmentation outputs. By incorporating SAM2's memory encoder and memory bank into a vision-language pipeline, Sa2VA achieves finer temporal alignment and better spatial precision across frames.

However, a structural bottleneck remains in the above line of work, including LISA, VISA, and Sa2VA, which all rely on a single `<SEG>` token for segmentation reasoning. Recent works such as VRS-HQ [Gong et al., 2025] and GLUS [Lin et al., 2025] have revisited the design of the special tokens. VRS-HQ introduces a temporal token (<TAK>) and a keyframe selection strategy to enhance temporal abstraction and occlusion-aware propagation, achieving strong results in long-form videos. Concurrently, GLUS [Lin et al., 2025] unifies global and local reasoning by employing a sparse set of context frames for semantic grounding and a continuous stream of query frames for object tracking, further strengthened by contrastive learning to diversify adjacency `<SEG>`s. Despite these improvements, both methods still fail to provide fine-grained frame-level segmentation supervision. To address this gap, we propose a multi-token reasoning framework that explicitly allocates distinct `<SEG>` tokens for different frames in a video. This design enables the model to retain detailed, object-specific cues across time while maintaining flexibility in language-driven reasoning.

## 3 Multi-Token Reasoning for Video Object Segmentation (MoVISA)

### 3.1 Problem Definition

**Video Object Segmentation.** Video object segmentation encompasses both *referring segmentation* and *reasoning segmentation*. Given an input video consisting of $T$ frames $I_{1:T} \in \mathbb{R}^{T \times H \times W \times 3}$ and a language expression $R$, the goal is to predict binary segmentation masks $\mathcal{M}_{1:T}$ of the referred object across all frames. Formally, the task can be formulated as learning a model $f_\theta$, with trainable parameters $\theta$, that predicts the mask sequence via

$$\mathcal{M}_{1:T} = f_\theta(I_{1:T}, R). \tag{1}$$

Video referring segmentation and video reasoning segmentation differ in the type of language expressions. Video referring segmentation typically uses direct prompts that describe the appearance or spatial location of target objects, while video reasoning segmentation focuses on indirect prompts that require common-sense reasoning.

**Dynamic Multi-Token Representation.** While prior methods such as Sa2VA [Yuan et al., 2025] formulate segmentation using a single `<SEG>` token per object, regardless of the video's temporal span, we introduce a dynamic multi-token segmentation strategy. Specifically, our approach assigns distinct `<SEG>` tokens for each object at each frame, enabling temporally dense supervision and explicit frame-level grounding.

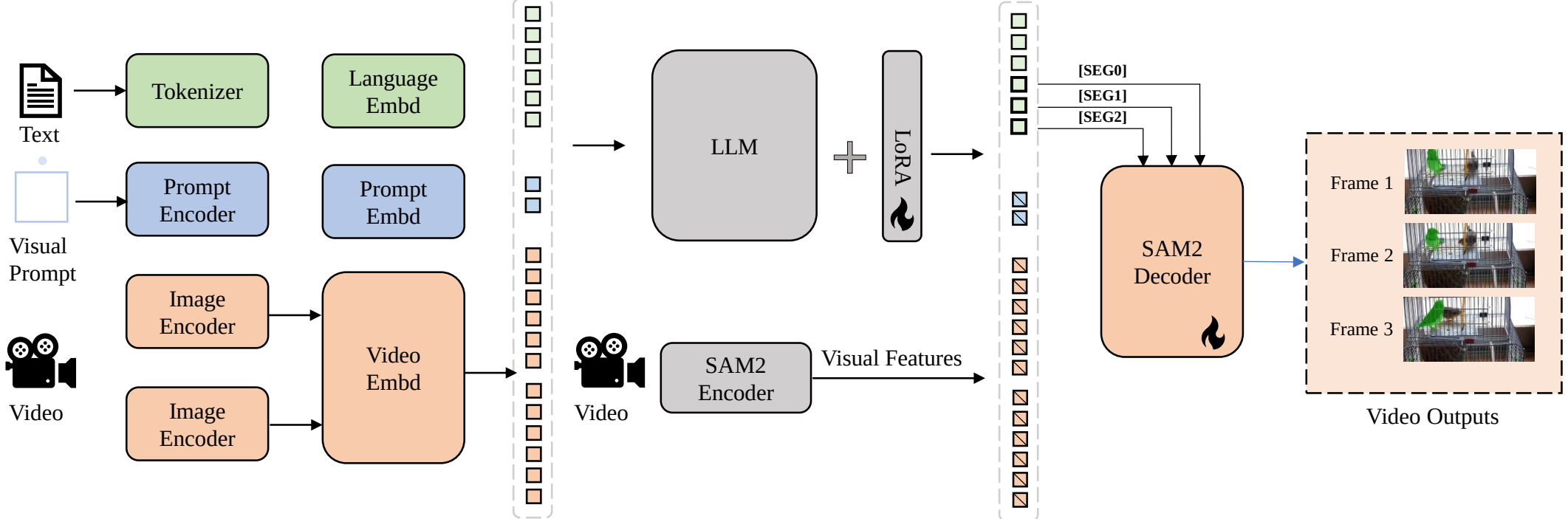

Figure 2: Overview of our network architecture. A Large Language Model (LLM) processes the input language and video embeddings and generates multiple segmentation tokens. These segmentation tokens are fed into the SAM2 mask decoder to generate the output segmentation.

**Unified Objective.** Let $R$ denote the tokenized language instruction, $V$ the visual input (either an image or video), and let $\{\texttt{<SEG>}_j\}_{j=1}^{K}$ refer to a set of learnable segmentation tokens, dynamically constructed based on the number of objects of interest and the number of input frames. The segmentation model $f_\theta$ predicts the target masks via

$$\mathcal{M} = f_\theta(R, V, \{\texttt{<SEG>}_j\}_{j=1}^{K}),$$

where $\mathcal{M}$ includes either spatial masks (for images) or spatio-temporal masklets (for videos). Although our primary focus is on video object segmentation, the flexibility of our `<SEG>` token design allows seamless adaptation to image segmentation by treating images as single-frame videos and employing a single `<SEG>` token.

### 3.2 Framework

As illustrated in Figure 2, our architecture follows a two-stream design, consisting of a pre-trained multi-modal LLM backbone and a SAM2 segmentation decoder. The framework is developed specifically for referring image and video object segmentation, and introduces a dynamic multi-token prompting mechanism that enables temporally grounded mask prediction at object-frame resolution.

#### 3.2.1 Visual-Language Backbone

We build our backbone on top of InternVL-2. Following Sa2VA, we adopt the same visual encoding and tokenization setup: each video frame is resized and compressed into 256 visual tokens using a vision encoder (e.g., CLIP or ViT). For video inputs, we sample a fixed number of key frames (typically 5), arrange them in temporal order, and encode them before the language instruction. For image inputs, we apply dynamic resolution encoding as proposed in InternVL-2 to preserve spatial fidelity.

The visual tokens pass through a projection layer and are combined with the language tokens. These are fed into the LLM to produce token outputs autoregressively. We extend the tokenizer to support custom special tokens, notably our indexed tokens $\{\texttt{<SEG>}_j\}_{j=1}^{K}$, which play a central role in conditioning the downstream segmentation.

#### 3.2.2 Special Token Design for Multi-Modal Reasoning

**Instruction Formulation via `<SEG>` Tokens.** We explore the strategy of incorporating segmentation tokens into the instruction-prompting framework, which relies on the generation ability of LLM, where the model is trained to autoregressively output segmentation tokens. In this setting, we structure the input as a multi-turn dialogue. The user sequentially queries the assistant to segment multiple objects, and the assistant responds by *generating* a list of segmentation tokens. Each object is associated with a unique block of <SEG> tokens: one per frame to capture spatial-temporal variation.

Formally, let the set of queried objects be $\mathcal{O} = \{o_1, o_2, \ldots, o_O\}$ and let the set of sampled frames be referred to be $\mathcal{T} = \{t_1, t_2, \ldots, t_T\}$. For each object $o_i \in \mathcal{O}$ and frame $t_j \in \mathcal{T}$, we define a corresponding segmentation token <SEG>$_{k_{i,m}}$, where:

$$k_{i,m} = (i-1) \cdot T + m, \quad m \in \{1, \ldots, T\}, \quad i \in \{1, \ldots, O\}.$$

The model is trained to autoregressively generate the structured response:

<user> Can you segment $o_i$ for $T$ frames?
<assistant> Sure, it is <SEG>$_{k_{1,1}}$, <SEG>$_{k_{1,2}}, \ldots,$ <SEG>$_{k_{O,T}}$.

This results in a set of $K = O \cdot T$ learnable segmentation tokens:

$$\mathcal{S} = \{\texttt{<SEG>}_0, \texttt{<SEG>}_1, \ldots, \texttt{<SEG>}_{K-1}\}.$$

Each token <SEG>$_k$ corresponds to an object-frame pair and serves as a spatial-temporal prompt for downstream mask decoding via a vision decoder (e.g., SAM2).

**Visual prompt injection via <vp> tokens.** To support object-level reasoning within a dialogue, we adopt the visual prompt understanding mechanism introduced in Sa2VA. For each referred object, we crop the corresponding image or frame region using its mask, extract the visual tokens from that region, and insert them into the language input stream. These cropped object features are flattened and wrapped with special tokens <vp> and </vp>, formatted as:

```
Object1:  <vp><IMG_FEAT>, ..., <IMG_FEAT></vp>.
```

Each object is assigned a symbolic identifier (e.g., Object1, Object2), which is used in the dialogue prompt. In subsequent dialogue turns, the model reasons about these object-specific features when generating <SEG> tokens. This strategy allows the LLM to maintain object identity across turns and resolve references explicitly within the conversation.

#### 3.2.3 Memory-Guided Mask Propagation

The hidden states corresponding to the predicted <SEG> tokens are extracted from the final layer of the LLM and projected via a two-layer MLP adapter into SAM2's prompt embedding space. These features serve as spatial-temporal mask prompts for the SAM2 decoder, similar to how box or click prompts are processed in SAM2 [Kirillov et al., 2023]. For each indexed token, SAM2 predicts a binary mask corresponding to the target object in its respective frame.

As in Sa2VA, we supervise only the masks at key frames and use the SAM2 memory encoder to propagate these predictions to non-key frames via memory attention. The key innovation in our method lies in using distinct <SEG> prompts per object and frame, which enriches the quality of key-frame segmentation and provides a more informative memory representation for propagation. In contrast, using the same segmentation across frames (as in Sa2VA [Yuan et al., 2025]) limits the temporal variability of the model.

#### 3.2.4 Unified Representation Across Modalities

While image and video inputs differ in temporal structure, our framework unifies prediction. Images are interpreted as one-frame videos, and accordingly, only one <SEG> token is generated per object. The same vision-language pipeline is applied, with token-level output supervision and mask generation via SAM2. This design enables seamless co-training on image and video referring segmentation tasks with minimal architectural changes.

### 3.3 Training

We train our model jointly on image and video object segmentation tasks using a one-shot instruction-tuning paradigm. Each training sample consists of a natural language query and either an image or a sampled video clip, with corresponding segmentation masks as supervision. We adopt a co-training setup across multiple datasets to promote generalization across spatial and spatio-temporal domains.

Table 1: Overview of training data and model size.

| Model | Backbone | Video Datasets (5.8K) | Image Datasets (56K) | Ref-SAV (37k) |
|---|---|---|---|---|
| MoVISA$^{\text{V}}$ | InternVL-2.5 4B | MeViS (2.0K), Ref-YTVOS (1.3K), ReVOS (2.5K) | - | - |
| MoVISA$^{\text{VIR}}$ | InternVL-2.0 1B | MeViS (2.0K), Ref-YTVOS (1.3K), ReVOS (2.5K) | RefCOCO, RefCOCO+, RefCOCOg (56K) | ✓ |

Our training objective consists of two components: (1) a language modeling loss for predicting the correct sequence of <SEG> tokens via the LLM, and (2) a segmentation loss applied to the masks produced by the SAM2 decoder. Following Sa2VA [Yuan et al., 2025], we define two losses: a cross-entropy loss $\mathcal{L}_{\text{text}}$ and a segmentation loss $\mathcal{L}_{\text{mask}}$. The language modeling loss $\mathcal{L}_{\text{text}}$ is the autoregressive cross-entropy loss for predicting the correct sequence of indexed <SEG> tokens:

$$\mathcal{L}_{\text{text}} = \text{CE}(\hat{y}_{\text{text}}, y_{\text{text}}).$$

Here, $y_{\text{text}}$ is the ground-truth sequence of <SEG> tokens, and $\hat{y}_{\text{text}}$ is the LLM-predicted sequence.

The segmentation loss $\mathcal{L}_{\text{mask}}$ is the sum of the pixel-wise binary cross-entropy loss and the Dice loss:

$$\mathcal{L}_{\text{mask}} = \mathcal{L}_{\text{CE}}(\hat{M}, M) + \mathcal{L}_{\text{DICE}}(\hat{M}, M).$$

$M, \hat{M}$ denote the ground truth and predicted masks respectively. To total loss for textitgeneration-based instruction is defined as:

$$\mathcal{L}_{\text{instruction}} = \mathcal{L}_{\text{text}} + \mathcal{L}_{\text{mask}}.$$

The model is trained end-to-end, and gradients are propagated both through the LLM (to improve <SEG> token prediction) and through the SAM2 decoder (to refine spatio-temporal segmentation performance).

### 3.4 Testing

During inference, the model processes referring instructions and corresponding visual inputs (image or video). The visual encoder extracts features from the input, which are projected to visual tokens and passed along with the query text into the LLM. The LLM autoregressively generates a sequence of uniquely indexed <SEG> tokens, corresponding to object-frame pairs.

Each predicted <SEG> token's hidden state is projected into the SAM2 prompt space via a learned adapter. For the key frames, these embeddings are used as prompts by SAM2 to generate segmentation masks. The generated masks are encoded into a memory bank via SAM2's memory encoder, and propagated to the non-key frames using memory attention.

## 4 Experiments

### 4.1 Training Data

We conduct our experiments under two configurations to accommodate computational constraints. The dataset sizes and corresponding model configurations are summarized in Table 1. First, we train the proposed MoVISA using the InternVL-2.5 4B backbone exclusively on video referring and reasoning segmentation datasets, including MeViS [Ding et al., 2023], Ref-YouTube-VOS [Oh et al., 2020], and ReVOS [Yan et al., 2024]. We refer to the resulting model as MoVISA$^{\text{V}}$(4B).

Second, due to limited computational resources, we train MoVISA with the lighter InternVL-2.0 1B backbone using a hybrid dataset setup. In this configuration, we combine the same video datasets (MeViS, Ref-YouTube-VOS, and ReVOS) with image segmentation datasets [Kazemzadeh et al., 2014, Yu et al., 2016] and additionally the Ref-SAV dataset proposed by [Yuan et al., 2025]. We use MoVISA$^{\text{VIR}}$(1B)to refer to the resulting model.

Table 2: Performance comparison on Referring VOS dataset.

| Methods | Backbone | | MeViS | Ref-YT-VOS | Ref-DAVIS17 |
|---|---|---|---|---|---|
| URVOS [Oh et al., 2020] | ResNet50 | 26M | 27.8 | 47.2 | 51.6 |
| LBDT [Ding et al., 2022] | ResNet50 | 26M | 29.3 | 49.4 | 54.1 |
| MTTR [Botach et al., 2022] | Video-Swin-T | 28M | 30.0 | 55.3 | – |
| ReferFormer [Wu et al., 2022] | Video-Swin-B | 50M | 31.0 | 62.9 | 61.1 |
| LMPM [Ding et al., 2023] | Swin-T | 28M | 37.2 | – | – |
| OnlineRefer [Wu et al., 2023] | Swin-L | 197M | – | 63.5 | 64.8 |
| LISA [Lai et al., 2024] | LLaVA | 13B | 37.9 | 54.4 | 66.0 |
| TrackGPT [Stroh, 2024] | LLaVA | 13B | 41.2 | 59.5 | 66.5 |
| VISA [Yan et al., 2024] | Chat-UniVi | 13B | 44.5 | 63.0 | 70.4 |
| LISA [Lai et al., 2024] | LLaVA | 7B | 37.2 | 53.9 | 64.8 |
| TrackGPT [Stroh, 2024] | LLaVA | 7B | 40.1 | 56.4 | 63.2 |
| VISA [Yan et al., 2024] | Chat-UniVi | 7B | 43.5 | 61.5 | 69.4 |
| GLUS$^{S}$ [Lin et al., 2025] | LISA-v1 | 7B | 50.3 | 66.6 | – |
| GLUS$^{A}$ [Lin et al., 2025] | LISA-v1 | 7B | 51.3 | 67.3 | – |
| Sa2VA-8B [Yuan et al., 2025] | InternVL-2.5 | 8B | 46.9 | 70.7 | 75.2 |
| Sa2VA-1B [Yuan et al., 2025] | InternVL-2.0 | 1B | 41.7 | 65.3 | 72.3 |
| MoVISA$^{VIR}$ (Ours) | InternVL-2.0 | 1B | 47.2 | 65.7 | 73.2 |
| Sa2VA-4B [Yuan et al., 2025] | InternVL-2.5 | 4B | 46.2 | 70.0 | 73.8 |
| MoVISA$^{V}$ (Ours) | InternVL-2.5 | 4B | 47.1 | 63.1 | 70.1 |

Table 3: Performance comparison on ReVOS dataset.

| Method | Backbone | | Referring | Reasoning | Overall |
|---|---|---|---|---|---|
| ReferFormer [Wu et al., 2022] | ResNet50 | 26M | 16.9 | 12.8 | 14.9 |
| MTTR [Botach et al., 2022] | Video-Swin-T | 50M | 30.0 | 21.0 | 25.5 |
| LMPM [Ding et al., 2023] | Swin-T | 28M | 34.1 | 18.8 | 26.4 |
| ReferFormer [Wu et al., 2022] | Video-Swin-B | 50M | 32.7 | 23.4 | 28.1 |
| LLaMA-VID [Li et al., 2024]+LMPM | Swin-T | 28M | 34.1 | 18.2 | 26.1 |
| VISA | Chat-UniVi | 13B | 54.1 | 40.9 | 47.5 |
| LISA [Lai et al., 2024] | LLaVA | 7B | 45.7 | 36.1 | 40.9 |
| VISA | Chat-UniVi | 7B | 52.9 | 39.2 | 46.1 |
| Sa2VA-8B [Yuan et al., 2025] | InternVL-2.5 | 8B | – | – | 57.6 |
| Sa2VA-1B [Yuan et al., 2025] | InternVL-2.0 | 1B | – | – | 47.6 |
| MoVISA$^{VIR}$ (Ours) | InternVL-2.0 | 1B | 56.8 | 46.4 | 51.6 |
| Sa2VA-4B [Yuan et al., 2025] | InternVL-2.5 | 4B | – | – | 53.2 |
| MoVISA$^{V}$ (Ours) | InternVL-2.5 | 4B | 59.1 | 47.5 | 53.3 |

### 4.2 Evaluation

We evaluate the model on a broad range of video referring segmentation benchmarks: Ref-YouTube-VOS [Oh et al., 2020], MeViS [Ding et al., 2023], Ref-DAVIS17 [Khoreva et al., 2018], and video reasoning segmentation benchmark: ReVOS [Yan et al., 2024]. We adopt the J&F metric, which combines region similarity (J) and contour accuracy (F), to assess segmentation quality.

### 4.3 Implementation Details

Our model is implemented using the XTuner [XTuner Contributors, 2023] codebase, which provides a unified infrastructure for multi-modal instruction tuning. During training, we freeze the visual encoder and fine-tune the LLM using LoRA [Hu et al., 2022], with an initial learning rate of $4\mathrm{e}{-5}$. The input sequence length is capped at 8,192 tokens to accommodate long-form visual-linguistic inputs across multiple frames.

Table 4: Ablation study on training datasets.

| Methods | Backbone | MeViS | Ref-YT-VOS | Ref-DAVIS17 |
|---|---|---|---|---|
| w/o image + Ref-SAV | InternVL-2.0 1B | 39.1 | 58.4 | 56.0 |
| MoVISA$^{VIR}$ (Ours) | InternVL-2.0 1B | 47.2 | 65.7 | 73.2 |

Table 5: Ablation of the number of segmentation tokens on Referring VOS data.

| Methods | Backbone | MeViS | Ref-YT-VOS | Ref-DAVIS17 |
|---|---|---|---|---|
| Single-Token | InternVL-2.0 1B | 31.0 | 46.0 | 56.0 |
| Frame Tokens | InternVL-2.0 1B | 30.9 | 46.6 | 55.0 |
| Frame-Object Tokens | InternVL-2.0 1B | 39.1 | 58.4 | 64.3 |

All experiments are conducted on six NVIDIA A6000 GPUs (48GB memory per GPU). Training with the InternVL-2.0 1B backbone on the video+image setup takes roughly 24 hours for 1 epoch, while training with the InternVL-2.5 4B backbone on the video-only setup takes approximately 7 hours for 2 epochs.

### 4.4 Main Results

**Referring VOS.** As shown in Tab. 2, compared with classic methods trained with significantly larger training datasets, our MoVISA$^{V}$(1B) still outperforms Sa2VA-4b by 0.9 J&F on the most challenging MEViS dataset. Our MoVISA$^{VIR}$ achieves a 5.5 point improvement compared to Sa2VA-1b while having a comparable model size and a similar training dataset size. On Ref-YT-VOS and Ref-DAVIS17 data, MoVISA$^{VIR}$ outperforms the Sa2VA-1b baseline by 0.4 J&F and 0.9 J&F respectively. However, we do observe a performance drop for MoVISA$^{V}$ on Ref-YT-VOS and Ref-DAVIS17, which highlights the benefits of adding image segmentation and the Ref-SAV data during training.

**Reason VOS.** In Tab. 3, we compare our proposed MoVISA model with existing state-of-art models using the ReVOS dataset. MoVISA$^{VIR}$(1B) outperforms Sa2VA-1B by 4.0 points on J&F, and MoVISA$^{VIR}$(4B) outperforms Sa2VA-4b by 0.1 points on J&F. Importantly, note that this is achieved while using significantly less training data, as Sa2VA-4b is trained with an additional 56K image data, and 37K Ref-SAV data.

### 4.5 Ablation Studies

**Training Datasets.** We first study the effectiveness of image segmentation data and the Ref-SAV dataset proposed by Sa2VA [Yuan et al., 2025]. In Tab. 4, we observe a significant performance drop across all datasets when using the InternVL-2.0 1B model as the backbone. The J&F scores drop by 8, 7.3, 17.2 points on MeViS, Ref-YT-VOS, and Ref-DAVIS17 respectively. This indicates that visual knowledge learned from image data is useful for training MLLM understanding tasks. Moreover, the descriptive content available in Ref-SAV data is also crucial for fine-grained visual concept understanding.

**Segmentation Token Design.** To demonstrate that our multi-token design for different frames and objects improves results, we compare to the following alternatives: (1) Single-Token: Following prior works such as LISA, VISA, and Sa2VA, we generate only a single `<SEG>` token for the entire video. (2) Frame Tokens: We generate indexed tokens `<SEG>`$_{k_T}$ for each frame $T$, but do not distinguish between different objects, thereby ignoring object variations during multi-turn dialogue formulation. (3) Frame-Object Tokens (Ours): Our proposed setting generates indexed tokens `<SEG>`$_{k_{o,t}}$ for each object $o$ and each frame $t$, explicitly modeling both object and frame variations.

From Tab. 5, we observe results to improve from single tokens to frame tokens to frame-object tokens, corroborating our claim that a single token is limiting.

**Decoder-Only Instruction** We follow the generation-based formulation in our main experiments, using the cross-entropy loss during training. To better understand the formulation, we also ablate whether a language-driven loss shifts the training objective away from object segmentation. To eliminate the reliance on language modeling supervision, we use a *decoder-only formulation*: all

Table 6: Decoder-only vs. Generate instruction ablation on Referring VOS dataset.

| Methods | Backbone | MeViS | Ref-YT-VOS | Ref-DAVIS17 |
|---|---|---|---|---|
| Decoder-only | InternVL-2.5 4B | 42.8 | – | 69.6 |
| Generate | InternVL-2.5 4B | 47.1 | 63.1 | 70.1 |

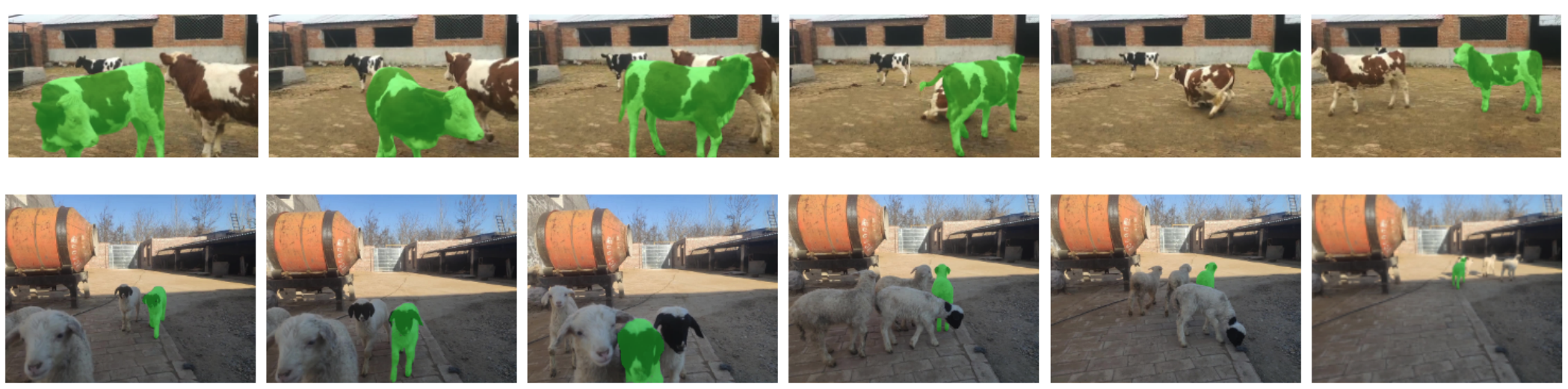

Figure 3: Qualitative Results. Prompts are "cow in the front first then moving to the right" (top row) and "lamb moving forward then turning around" (bottom row).

<SEG> tokens are inserted into the input prompt. Rather than training the model to generate these tokens, we extract their hidden states directly from the decoder output and use them as grounding signals for segmentation. Such, we have total loss as $\mathcal{L}_{\text{instruction}} = \mathcal{L}_{\text{mask}}$. The input prompt remains structurally identical to the original formulation:

<user>: Can you segment $o_i$ for $T$ frames? Sure, it is
<SEG>$_{k_{1,1}}$, <SEG>$_{k_{1,2}}$, ..., <SEG>$_{k_{O,T}}$.
<assistant>: [free_text]

However, since all $K = O \cdot T$ segmentation tokens are already provided as input, we don't supervise their generation. Therefore, we remove the cross-entropy loss on language outputs and supervise the model only through the predicted segmentation masks derived from the <SEG> tokens' hidden states. We find that this formulation simplifies training and shifts the model's learning focus from syntactic token prediction to segmentation. From Tab. 6, we observe the generate approach to yield better results.

**Qualitative Results.** We visualize results in Fig. 3. Our MoVISA$^{\text{VIR}}$ successfully captures the object referred to via the prompt "cow in the front first then moving to the right" and "lamb moving forward then turning around" (MEVIS data) . Our multi-token design provides sufficient details to handle ambiguous objects with varying positions and multiple instances in the video.

## 5 Conclusion

The single <SEG> token design of recent video object segmentation methods with multimodal large language models fundamentally limits the representational capacity. Corresponding methods struggle to represent objects which change significantly across the frames of a given video. To address this limitation we study a multi-token design, i.e., "Multi-Token Reasoning for Video Object Segmentation" (MoVISA). We find MoVISA to yield compelling results on challenging datasets.

**Limitations and Broader Impact.** Reasoning-based video object segmentation is an important task for embodied AI and virtual assistants. The broader impacts of improved methods are hence significant. E.g., it can be used to improve environment understanding of autonomous robots which interact with the real-world. Alternatively, it can aid the elderly and the disabled to navigate environments more comfortably. However, reasoning-based video object segmentation technology can also be abused for mass-surveillance or unwanted tracking. Every deployment differs and should be implemented with great care.

Due to computational limitations, in this work, we focus on 1B and 4B parameter models. While we think these models are particularly useful for real-world use-cases, it is certainly valuable to assess improvements for larger models, which is beyond our capabilities.

## References

Jean-Baptiste Alayrac, Jeff Donahue, Pauline Luc, Antoine Miech, Iain Barr, Yana Hasson, Karel Lenc, Arthur Mensch, Katie Millican, Malcolm Reynolds, Roman Ring, Eliza Rutherford, Serkan Cabi, Tengda Han, Zhitao Gong, Sina Samangooei, Marianne Monteiro, Jacob Menick, Sebastian Borgeaud, Andrew Brock, Aida Nematzadeh, Sahand Sharifzadeh, Mikolaj Binkowski, Ricardo Barreira, Oriol Vinyals, Andrew Zisserman, and Karen Simonyan. Flamingo: a visual language model for few-shot learning. *NeurIPS*, 2022.

Adam Botach, Evgenii Zheltonozhskii, and Chaim Baskin. End-to-end referring video object segmentation with multimodal transformers. *CVPR*, 2022.

Henghui Ding, Chang Liu, Shuting He, Xudong Jiang, and Chen Change Loy. Mevis: A large-scale benchmark for video segmentation with motion expressions. *ICCV*, 2023.

Zihan Ding, Tianrui Hui, Junshi Huang, Xiaoming Wei, Jizhong Han, and Si Liu. Language-bridged spatial-temporal interaction for referring video object segmentation. *CVPR*, 2022.

Sitong Gong, Yunzhi Zhuge, Lu Zhang, Zongxin Yang, Pingping Zhang, and Huchuan Lu. The devil is in temporal token: High quality video reasoning segmentation. *CVPR*, 2025.

Edward J. Hu, Yelong Shen, Phillip Wallis, Zeyuan Allen-Zhu, Yuanzhi Li, Shean Wang, Lu Wang, and Weizhu Chen. Lora: Low-rank adaptation of large language models. *ICLR*, 2022.

Sahar Kazemzadeh, Vicente Ordonez, Mark Matten, and Tamara Berg. ReferItGame: Referring to objects in photographs of natural scenes. In *EMNLP*, 2014.

Anna Khoreva, Anna Rohrbach, and Bernt Schiele. Video object segmentation with language referring expressions. *ACCV*, 2018.

Alexander Kirillov, Eric Mintun, Nikhila Ravi, Hanzi Mao, Chloe Rolland, Laura Gustafson, Tete Xiao, Spencer Whitehead, Alexander C. Berg, Wan-Yen Lo, Piotr Dollár, and Ross Girshick. Segment anything. *ICCV*, 2023.

Xin Lai, Zhuotao Tian, Yukang Chen, Yanwei Li, Yuhui Yuan, Shu Liu, and Jiaya Jia. Lisa: Reasoning segmentation via large language model. *CVPR*, 2024.

Bo Li, Yuanhan Zhang, Liangyu Chen, Jinghao Wang, Jingkang Yang, and Ziwei Liu. Otter: A multi-modal model with in-context instruction tuning, 2023a. URL `https://arxiv.org/abs/2305.03726`.

Junnan Li, Dongxu Li, Silvio Savarese, and Steven Hoi. Blip-2: Bootstrapping language-image pre-training with frozen image encoders and large language models. *ICML*, 2023b.

Yanwei Li, Chengyao Wang, and Jiaya Jia. Llama-vid: An image is worth 2 tokens in large language models. *ECCV*, 2024.

Bin Lin, Yang Ye, Bin Zhu, Jiaxi Cui, Munan Ning, Peng Jin, and Li Yuan. Video-llava: Learning united visual representation by alignment before projection. *EMNLP*, 2024.

Lang Lin, Xueyang Yu, Ziqi Pang, and Yu-Xiong Wang. Glus: Global-local reasoning unified into a single large language model for video segmentation. *CVPR*, 2025.

Haotian Liu, Chunyuan Li, Qingyang Wu, and Yong Jae Lee. Visual instruction tuning. *NeurIPS*, 2023.

Muhammad Maaz, Hanoona Rasheed, Salman Khan, and Fahad Shahbaz Khan. Video-chatgpt: Towards detailed video understanding via large vision and language models. *ACL*, 2024.

Seoung Wug Oh, Joon-Young Lee, and Bohyung Han. Urvos: Unified referring video object segmentation network with a large-scale benchmark. In *ECCV*, 2020.

OpenAI, Josh Achiam, Steven Adler, Sandhini Agarwal, Lama Ahmad, Ilge Akkaya, Florencia Leoni Aleman, Diogo Almeida, Janko Altenschmidt, Sam Altman, Shyamal Anadkat, Red Avila, Igor Babuschkin, Suchir Balaji, Valerie Balcom, Paul Baltescu, Haiming Bao, Mohammad Bavarian, Jeff Belgum, Irwan Bello, Jake Berdine, Gabriel Bernadett-Shapiro, Christopher Berner, Lenny Bogdonoff, Oleg Boiko, Madelaine Boyd, Anna-Luisa Brakman, Greg Brockman, Tim Brooks, Miles Brundage, Kevin Button, Trevor Cai, Rosie Campbell, Andrew Cann, Brittany Carey, Chelsea Carlson, Rory Carmichael, Brooke Chan, Che Chang, Fotis Chantzis, Derek Chen, Sully Chen, Ruby Chen, Jason Chen, Mark Chen, Ben Chess, Chester Cho, Casey Chu, Hyung Won Chung, Dave Cummings, Jeremiah Currier, Yunxing Dai, Cory Decareaux, Thomas Degry, Noah Deutsch, Damien Deville, Arka Dhar, David Dohan, Steve Dowling, Sheila Dunning, Adrien Ecoffet, Atty Eleti, Tyna Eloundou, David Farhi, Liam Fedus, Niko Felix, Simón Posada Fishman, Juston Forte, Isabella Fulford, Leo Gao, Elie Georges, Christian Gibson, Vik Goel, Tarun Gogineni, Gabriel Goh, Rapha Gontijo-Lopes, Jonathan Gordon, Morgan Grafstein, Scott Gray, Ryan Greene, Joshua Gross, Shixiang Shane Gu, Yufei Guo, Chris Hallacy, Jesse Han, Jeff Harris, Yuchen He, Mike Heaton, Johannes Heidecke, Chris Hesse, Alan Hickey, Wade Hickey, Peter Hoeschele, Brandon Houghton, Kenny Hsu, Shengli Hu, Xin Hu, Joost Huizinga, Shantanu Jain, Shawn Jain, Joanne Jang, Angela Jiang, Roger Jiang, Haozhun Jin, Denny Jin, Shino Jomoto, Billie Jonn, Heewoo Jun, Tomer Kaftan, Łukasz Kaiser, Ali Kamali, Ingmar Kanitscheider, Nitish Shirish Keskar, Tabarak Khan, Logan Kilpatrick, Jong Wook Kim, Christina Kim, Yongjik Kim, Jan Hendrik Kirchner, Jamie Kiros, Matt Knight, Daniel Kokotajlo, Łukasz Kondraciuk, Andrew Kondrich, Aris Konstantinidis, Kyle Kosic, Gretchen Krueger, Vishal Kuo, Michael Lampe, Ikai Lan, Teddy Lee, Jan Leike, Jade Leung, Daniel Levy, Chak Ming Li, Rachel Lim, Molly Lin, Stephanie Lin, Mateusz Litwin, Theresa Lopez, Ryan Lowe, Patricia Lue, Anna Makanju, Kim Malfacini, Sam Manning, Todor Markov, Yaniv Markovski, Bianca Martin, Katie Mayer, Andrew Mayne, Bob McGrew, Scott Mayer McKinney, Christine McLeavey, Paul McMillan, Jake McNeil, David Medina, Aalok Mehta, Jacob Menick, Luke Metz, Andrey Mishchenko, Pamela Mishkin, Vinnie Monaco, Evan Morikawa, Daniel Mossing, Tong Mu, Mira Murati, Oleg Murk, David Mély, Ashvin Nair, Reiichiro Nakano, Rajeev Nayak, Arvind Neelakantan, Richard Ngo, Hyeonwoo Noh, Long Ouyang, Cullen O'Keefe, Jakub Pachocki, Alex Paino, Joe Palermo, Ashley Pantuliano, Giambattista Parascandolo, Joel Parish, Emy Parparita, Alex Passos, Mikhail Pavlov, Andrew Peng, Adam Perelman, Filipe de Avila Belbute Peres, Michael Petrov, Henrique Ponde de Oliveira Pinto, Michael, Pokorny, Michelle Pokrass, Vitchyr H. Pong, Tolly Powell, Alethea Power, Boris Power, Elizabeth Proehl, Raul Puri, Alec Radford, Jack Rae, Aditya Ramesh, Cameron Raymond, Francis Real, Kendra Rimbach, Carl Ross, Bob Rotsted, Henri Roussez, Nick Ryder, Mario Saltarelli, Ted Sanders, Shibani Santurkar, Girish Sastry, Heather Schmidt, David Schnurr, John Schulman, Daniel Selsam, Kyla Sheppard, Toki Sherbakov, Jessica Shieh, Sarah Shoker, Pranav Shyam, Szymon Sidor, Eric Sigler, Maddie Simens, Jordan Sitkin, Katarina Slama, Ian Sohl, Benjamin Sokolowsky, Yang Song, Natalie Staudacher, Felipe Petroski Such, Natalie Summers, Ilya Sutskever, Jie Tang, Nikolas Tezak, Madeleine B. Thompson, Phil Tillet, Amin Tootoonchian, Elizabeth Tseng, Preston Tuggle, Nick Turley, Jerry Tworek, Juan Felipe Cerón Uribe, Andrea Vallone, Arun Vijayvergiya, Chelsea Voss, Carroll Wainwright, Justin Jay Wang, Alvin Wang, Ben Wang, Jonathan Ward, Jason Wei, CJ Weinmann, Akila Welihinda, Peter Welinder, Jiayi Weng, Lilian Weng, Matt Wiethoff, Dave Willner, Clemens Winter, Samuel Wolrich, Hannah Wong, Lauren Workman, Sherwin Wu, Jeff Wu, Michael Wu, Kai Xiao, Tao Xu, Sarah Yoo, Kevin Yu, Qiming Yuan, Wojciech Zaremba, Rowan Zellers, Chong Zhang, Marvin Zhang, Shengjia Zhao, Tianhao Zheng, Juntang Zhuang, William Zhuk, and Barret Zoph. Gpt-4 technical report, 2024. URL `https://arxiv.org/abs/2303.08774`.

Nikhila Ravi, Valentin Gabeur, Yuan-Ting Hu, Ronghang Hu, Chaitanya Ryali, Tengyu Ma, Haitham Khedr, Roman Rädle, Chloe Rolland, Laura Gustafson, Eric Mintun, Junting Pan, Kalyan Vasudev Alwala, Nicolas Carion, Chao-Yuan Wu, Ross Girshick, Piotr Dollár, and Christoph Feichtenhofer. Sam 2: Segment anything in images and videos. *ICLR*, 2025.

Nicholas Stroh. Trackgpt – a generative pre-trained transformer for cross-domain entity trajectory forecasting, 2024. URL `https://arxiv.org/abs/2402.00066`.

Dongming Wu, Tiancai Wang, Yuang Zhang, Xiangyu Zhang, and Jianbing Shen. Onlinerefer: A simple online baseline for referring video object segmentation. *ICCV*, 2023.

Jiannan Wu, Yi Jiang, Peize Sun, Zehuan Yuan, and Ping Luo. Language as queries for referring video object segmentation. *CVPR*, 2022.

XTuner Contributors. XTuner: A Toolkit for Efficiently Fine-Tuning LLM. https://github.com/InternLM/xtuner, 2023. Accessed: 2025-05-12.

Bin Yan, Yi Jiang, Jiannan Wu, Dong Wang, Ping Luo, Zehuan Yuan, and Huchuan Lu. Universal instance perception as object discovery and retrieval. *CVPR*, 2023.

Cilin Yan, Haochen Wang, Shilin Yan, Xiaolong Jiang, Yao Hu, Guoliang Kang, Weidi Xie, and Efstratios Gavves. Visa: Reasoning video object segmentation via large language models. *ECCV*, 2024.

Qinghao Ye, Haiyang Xu, Guohai Xu, Jiabo Ye, Ming Yan, Yiyang Zhou, Junyang Wang, Anwen Hu, Pengcheng Shi, Yaya Shi, Chenliang Li, Yuanhong Xu, Hehong Chen, Junfeng Tian, Qi Qian, Ji Zhang, Fei Huang, and Jingren Zhou. mplug-owl: Modularization empowers large language models with multimodality, 2024. URL https://arxiv.org/abs/2304.14178.

Licheng Yu, Patrick Poirson, Shan Yang, Alexander C. Berg, and Tamara L. Berg. Modeling context in referring expressions. *ECCV*, 2016.

Licheng Yu, Zhe Lin, Xiaohui Shen, Jimei Yang, Xin Lu, Mohit Bansal, and Tamara L. Berg. Mattnet: Modular attention network for referring expression comprehension. *CVPR*, 2018.

Haobo Yuan, Xiangtai Li, Tao Zhang, Zilong Huang, Shilin Xu, Shunping Ji, Yunhai Tong, Lu Qi, Jiashi Feng, and Ming-Hsuan Yang. Sa2va: Marrying sam2 with llava for dense grounded understanding of images and videos, 2025. URL https://arxiv.org/abs/2501.04001.

Deyao Zhu, Jun Chen, Xiaoqian Shen, Xiang Li, and Mohamed Elhoseiny. Minigpt-4: Enhancing vision-language understanding with advanced large language models, 2023. URL https://arxiv.org/abs/2304.10592.

## NeurIPS Paper Checklist

1. **Claims**

   Question: Do the main claims made in the abstract and introduction accurately reflect the paper's contributions and scope?

   Answer: [Yes]

   Justification: The abstract and the introduction accurately reflect the contribution and scope of this paper.

   Guidelines:

   - The answer NA means that the abstract and introduction do not include the claims made in the paper.
   - The abstract and/or introduction should clearly state the claims made, including the contributions made in the paper and important assumptions and limitations. A No or NA answer to this question will not be perceived well by the reviewers.
   - The claims made should match theoretical and experimental results, and reflect how much the results can be expected to generalize to other settings.
   - It is fine to include aspirational goals as motivation as long as it is clear that these goals are not attained by the paper.

2. **Limitations**

   Question: Does the paper discuss the limitations of the work performed by the authors?

   Answer: [Yes]

   Justification: We discuss the limitations of our method in Section 5.

   Guidelines:

   - The answer NA means that the paper has no limitation while the answer No means that the paper has limitations, but those are not discussed in the paper.
   - The authors are encouraged to create a separate "Limitations" section in their paper.
   - The paper should point out any strong assumptions and how robust the results are to violations of these assumptions (e.g., independence assumptions, noiseless settings, model well-specification, asymptotic approximations only holding locally). The authors should reflect on how these assumptions might be violated in practice and what the implications would be.
   - The authors should reflect on the scope of the claims made, e.g., if the approach was only tested on a few datasets or with a few runs. In general, empirical results often depend on implicit assumptions, which should be articulated.
   - The authors should reflect on the factors that influence the performance of the approach. For example, a facial recognition algorithm may perform poorly when image resolution is low or images are taken in low lighting. Or a speech-to-text system might not be used reliably to provide closed captions for online lectures because it fails to handle technical jargon.
   - The authors should discuss the computational efficiency of the proposed algorithms and how they scale with dataset size.
   - If applicable, the authors should discuss possible limitations of their approach to address problems of privacy and fairness.
   - While the authors might fear that complete honesty about limitations might be used by reviewers as grounds for rejection, a worse outcome might be that reviewers discover limitations that aren't acknowledged in the paper. The authors should use their best judgment and recognize that individual actions in favor of transparency play an important role in developing norms that preserve the integrity of the community. Reviewers will be specifically instructed to not penalize honesty concerning limitations.

3. **Theory assumptions and proofs**

   Question: For each theoretical result, does the paper provide the full set of assumptions and a complete (and correct) proof?

   Answer: [NA]

Justification: This paper does not include theoretical results.

Guidelines:

- The answer NA means that the paper does not include theoretical results.
- All the theorems, formulas, and proofs in the paper should be numbered and cross-referenced.
- All assumptions should be clearly stated or referenced in the statement of any theorems.
- The proofs can either appear in the main paper or the supplemental material, but if they appear in the supplemental material, the authors are encouraged to provide a short proof sketch to provide intuition.
- Inversely, any informal proof provided in the core of the paper should be complemented by formal proofs provided in appendix or supplemental material.
- Theorems and Lemmas that the proof relies upon should be properly referenced.

4. **Experimental result reproducibility**

   Question: Does the paper fully disclose all the information needed to reproduce the main experimental results of the paper to the extent that it affects the main claims and/or conclusions of the paper (regardless of whether the code and data are provided or not)?

   Answer: [Yes]

   Justification: We describe our method with sufficient details to reproduce the main experimental results. We will also provide open-source code for reproducibility.

   Guidelines:

   - The answer NA means that the paper does not include experiments.
   - If the paper includes experiments, a No answer to this question will not be perceived well by the reviewers: Making the paper reproducible is important, regardless of whether the code and data are provided or not.
   - If the contribution is a dataset and/or model, the authors should describe the steps taken to make their results reproducible or verifiable.
   - Depending on the contribution, reproducibility can be accomplished in various ways. For example, if the contribution is a novel architecture, describing the architecture fully might suffice, or if the contribution is a specific model and empirical evaluation, it may be necessary to either make it possible for others to replicate the model with the same dataset, or provide access to the model. In general. releasing code and data is often one good way to accomplish this, but reproducibility can also be provided via detailed instructions for how to replicate the results, access to a hosted model (e.g., in the case of a large language model), releasing of a model checkpoint, or other means that are appropriate to the research performed.
   - While NeurIPS does not require releasing code, the conference does require all submissions to provide some reasonable avenue for reproducibility, which may depend on the nature of the contribution. For example
     (a) If the contribution is primarily a new algorithm, the paper should make it clear how to reproduce that algorithm.
     (b) If the contribution is primarily a new model architecture, the paper should describe the architecture clearly and fully.
     (c) If the contribution is a new model (e.g., a large language model), then there should either be a way to access this model for reproducing the results or a way to reproduce the model (e.g., with an open-source dataset or instructions for how to construct the dataset).
     (d) We recognize that reproducibility may be tricky in some cases, in which case authors are welcome to describe the particular way they provide for reproducibility. In the case of closed-source models, it may be that access to the model is limited in some way (e.g., to registered users), but it should be possible for other researchers to have some path to reproducing or verifying the results.

5. **Open access to data and code**

   Question: Does the paper provide open access to the data and code, with sufficient instructions to faithfully reproduce the main experimental results, as described in supplemental material?

Answer: [No]

Justification: We'll release code to ensure reproducibility.

Guidelines:

- The answer NA means that paper does not include experiments requiring code.
- Please see the NeurIPS code and data submission guidelines (`https://nips.cc/public/guides/CodeSubmissionPolicy`) for more details.
- While we encourage the release of code and data, we understand that this might not be possible, so "No" is an acceptable answer. Papers cannot be rejected simply for not including code, unless this is central to the contribution (e.g., for a new open-source benchmark).
- The instructions should contain the exact command and environment needed to run to reproduce the results. See the NeurIPS code and data submission guidelines (`https://nips.cc/public/guides/CodeSubmissionPolicy`) for more details.
- The authors should provide instructions on data access and preparation, including how to access the raw data, preprocessed data, intermediate data, and generated data, etc.
- The authors should provide scripts to reproduce all experimental results for the new proposed method and baselines. If only a subset of experiments are reproducible, they should state which ones are omitted from the script and why.
- At submission time, to preserve anonymity, the authors should release anonymized versions (if applicable).
- Providing as much information as possible in supplemental material (appended to the paper) is recommended, but including URLs to data and code is permitted.

6. **Experimental setting/details**

   Question: Does the paper specify all the training and test details (e.g., data splits, hyperparameters, how they were chosen, type of optimizer, etc.) necessary to understand the results?

   Answer: [Yes]

   Justification: We specify all necessary experimental and implementation details.

   Guidelines:

   - The answer NA means that the paper does not include experiments.
   - The experimental setting should be presented in the core of the paper to a level of detail that is necessary to appreciate the results and make sense of them.
   - The full details can be provided either with the code, in appendix, or as supplemental material.

7. **Experiment statistical significance**

   Question: Does the paper report error bars suitably and correctly defined or other appropriate information about the statistical significance of the experiments?

   Answer: [No]

   Justification: We do not report statistical significance as the experiments are costly, in line with prior works.

   Guidelines:

   - The answer NA means that the paper does not include experiments.
   - The authors should answer "Yes" if the results are accompanied by error bars, confidence intervals, or statistical significance tests, at least for the experiments that support the main claims of the paper.
   - The factors of variability that the error bars are capturing should be clearly stated (for example, train/test split, initialization, random drawing of some parameter, or overall run with given experimental conditions).
   - The method for calculating the error bars should be explained (closed form formula, call to a library function, bootstrap, etc.)
   - The assumptions made should be given (e.g., Normally distributed errors).

- It should be clear whether the error bar is the standard deviation or the standard error of the mean.
- It is OK to report 1-sigma error bars, but one should state it. The authors should preferably report a 2-sigma error bar than state that they have a 96% CI, if the hypothesis of Normality of errors is not verified.
- For asymmetric distributions, the authors should be careful not to show in tables or figures symmetric error bars that would yield results that are out of range (e.g. negative error rates).
- If error bars are reported in tables or plots, The authors should explain in the text how they were calculated and reference the corresponding figures or tables in the text.

8. **Experiments compute resources**

   Question: For each experiment, does the paper provide sufficient information on the computer resources (type of compute workers, memory, time of execution) needed to reproduce the experiments?

   Answer: [Yes]

   Justification: Yes, we provide the details in Section 4.3.

   Guidelines:

   - The answer NA means that the paper does not include experiments.
   - The paper should indicate the type of compute workers CPU or GPU, internal cluster, or cloud provider, including relevant memory and storage.
   - The paper should provide the amount of compute required for each of the individual experimental runs as well as estimate the total compute.
   - The paper should disclose whether the full research project required more compute than the experiments reported in the paper (e.g., preliminary or failed experiments that didn't make it into the paper).

9. **Code of ethics**

   Question: Does the research conducted in the paper conform, in every respect, with the NeurIPS Code of Ethics `https://neurips.cc/public/EthicsGuidelines`?

   Answer: [Yes]

   Justification: Our research conform, in every respect, with the NeurIPS Code of Ethics.

   Guidelines:

   - The answer NA means that the authors have not reviewed the NeurIPS Code of Ethics.
   - If the authors answer No, they should explain the special circumstances that require a deviation from the Code of Ethics.
   - The authors should make sure to preserve anonymity (e.g., if there is a special consideration due to laws or regulations in their jurisdiction).

10. **Broader impacts**

    Question: Does the paper discuss both potential positive societal impacts and negative societal impacts of the work performed?

    Answer: [NA]

    Justification: Our research is foundational research in computer vision, with no direct societal impact.

    Guidelines:

    - The answer NA means that there is no societal impact of the work performed.
    - If the authors answer NA or No, they should explain why their work has no societal impact or why the paper does not address societal impact.
    - Examples of negative societal impacts include potential malicious or unintended uses (e.g., disinformation, generating fake profiles, surveillance), fairness considerations (e.g., deployment of technologies that could make decisions that unfairly impact specific groups), privacy considerations, and security considerations.

- The conference expects that many papers will be foundational research and not tied to particular applications, let alone deployments. However, if there is a direct path to any negative applications, the authors should point it out. For example, it is legitimate to point out that an improvement in the quality of generative models could be used to generate deepfakes for disinformation. On the other hand, it is not needed to point out that a generic algorithm for optimizing neural networks could enable people to train models that generate Deepfakes faster.
- The authors should consider possible harms that could arise when the technology is being used as intended and functioning correctly, harms that could arise when the technology is being used as intended but gives incorrect results, and harms following from (intentional or unintentional) misuse of the technology.
- If there are negative societal impacts, the authors could also discuss possible mitigation strategies (e.g., gated release of models, providing defenses in addition to attacks, mechanisms for monitoring misuse, mechanisms to monitor how a system learns from feedback over time, improving the efficiency and accessibility of ML).

11. **Safeguards**

    Question: Does the paper describe safeguards that have been put in place for responsible release of data or models that have a high risk for misuse (e.g., pretrained language models, image generators, or scraped datasets)?

    Answer: [NA]

    Justification: This paper poses no such risks.

    Guidelines:

    - The answer NA means that the paper poses no such risks.
    - Released models that have a high risk for misuse or dual-use should be released with necessary safeguards to allow for controlled use of the model, for example by requiring that users adhere to usage guidelines or restrictions to access the model or implementing safety filters.
    - Datasets that have been scraped from the Internet could pose safety risks. The authors should describe how they avoided releasing unsafe images.
    - We recognize that providing effective safeguards is challenging, and many papers do not require this, but we encourage authors to take this into account and make a best faith effort.

12. **Licenses for existing assets**

    Question: Are the creators or original owners of assets (e.g., code, data, models), used in the paper, properly credited and are the license and terms of use explicitly mentioned and properly respected?

    Answer: [Yes]

    Justification: We cite the papers that release the assets. Specific licenses can be obtained from the authors of those papers.

    Guidelines:

    - The answer NA means that the paper does not use existing assets.
    - The authors should cite the original paper that produced the code package or dataset.
    - The authors should state which version of the asset is used and, if possible, include a URL.
    - The name of the license (e.g., CC-BY 4.0) should be included for each asset.
    - For scraped data from a particular source (e.g., website), the copyright and terms of service of that source should be provided.
    - If assets are released, the license, copyright information, and terms of use in the package should be provided. For popular datasets, `paperswithcode.com/datasets` has curated licenses for some datasets. Their licensing guide can help determine the license of a dataset.
    - For existing datasets that are re-packaged, both the original license and the license of the derived asset (if it has changed) should be provided.

- If this information is not available online, the authors are encouraged to reach out to the asset's creators.

13. **New assets**

    Question: Are new assets introduced in the paper well documented and is the documentation provided alongside the assets?

    Answer: [NA]

    Justification: This paper does not release new assets.

    Guidelines:

    - The answer NA means that the paper does not release new assets.
    - Researchers should communicate the details of the dataset/code/model as part of their submissions via structured templates. This includes details about training, license, limitations, etc.
    - The paper should discuss whether and how consent was obtained from people whose asset is used.
    - At submission time, remember to anonymize your assets (if applicable). You can either create an anonymized URL or include an anonymized zip file.

14. **Crowdsourcing and research with human subjects**

    Question: For crowdsourcing experiments and research with human subjects, does the paper include the full text of instructions given to participants and screenshots, if applicable, as well as details about compensation (if any)?

    Answer: [NA]

    Justification: This paper does not involve crowdsourcing nor research with human subjects.

    Guidelines:

    - The answer NA means that the paper does not involve crowdsourcing nor research with human subjects.
    - Including this information in the supplemental material is fine, but if the main contribution of the paper involves human subjects, then as much detail as possible should be included in the main paper.
    - According to the NeurIPS Code of Ethics, workers involved in data collection, curation, or other labor should be paid at least the minimum wage in the country of the data collector.

15. **Institutional review board (IRB) approvals or equivalent for research with human subjects**

    Question: Does the paper describe potential risks incurred by study participants, whether such risks were disclosed to the subjects, and whether Institutional Review Board (IRB) approvals (or an equivalent approval/review based on the requirements of your country or institution) were obtained?

    Answer: [NA]

    Justification: This paper does not involve crowdsourcing nor research with human subjects.

    Guidelines:

    - The answer NA means that the paper does not involve crowdsourcing nor research with human subjects.
    - Depending on the country in which research is conducted, IRB approval (or equivalent) may be required for any human subjects research. If you obtained IRB approval, you should clearly state this in the paper.
    - We recognize that the procedures for this may vary significantly between institutions and locations, and we expect authors to adhere to the NeurIPS Code of Ethics and the guidelines for their institution.
    - For initial submissions, do not include any information that would break anonymity (if applicable), such as the institution conducting the review.

16. **Declaration of LLM usage**

Question: Does the paper describe the usage of LLMs if it is an important, original, or non-standard component of the core methods in this research? Note that if the LLM is used only for writing, editing, or formatting purposes and does not impact the core methodology, scientific rigorousness, or originality of the research, declaration is not required.

Answer: [NA]

Justification: The core method development in this research does not involve LLMs as any important, original, or non-standard components.

Guidelines:

- The answer NA means that the core method development in this research does not involve LLMs as any important, original, or non-standard components.
- Please refer to our LLM policy (https://neurips.cc/Conferences/2025/LLM) for what should or should not be described.